\documentclass[10pt]{cai}

\graphicspath{{figs/}}

\begin{document}

\volumeheader{34}{\href{https://doi.org/10.21428/594757db.d2f8342b}{10.21428/594757db.d2f8342b}}
\begin{center}

  \title{Eye-focused Detection of Bell's Palsy in Videos}
  \maketitle

  \thispagestyle{empty}

  \begin{tabular}{cc}
    Sharik Ali Ansari\upstairs{\affilone}, Koteswar Rao Jerripothula\upstairs{\affiltwo,*}, Pragya Nagpal\upstairs{\affilthree}, Ankush Mittal\upstairs{\affilfour}
   \\[0.25ex]
   {\small \upstairs{\affilone} College of Engineering Roorkee, India} \\
   {\small \upstairs{\affiltwo} Indraprastha Institute of Information Technology Delhi (IIIT-Delhi), India} \\
   {\small \upstairs{\affilthree} University of Texas at Dallas, USA} \\
   {\small \upstairs{\affilfour} Raman Classes, Rookee, India} \\
  \end{tabular}
  
  \emails{
    \upstairs{*}koteswar@iiitd.ac.in \\
    Published in the Proceedings of the $34^{th}$ Canadian Conference on Artificial Intelligence. Please cite \cite{Ansari2021Eye} when referencing this paper}
  \vspace*{0.2in}
\end{center}

\begin{abstract}

In this paper, we present how Bell's Palsy, a neurological disorder, can be detected just from a subject's eyes in a video. We notice that Bell's Palsy patients often struggle to blink their eyes on the affected side. As a result, we can observe a clear contrast between the blinking patterns of the two eyes. Although previous works did utilize images/videos to detect this disorder, none have explicitly focused on the eyes. Most of them require the entire face. One obvious advantage of having an eye-focused detection system is that subjects' anonymity is not at risk. Also, our AI decisions based on simple blinking patterns make them explainable and straightforward. Specifically, we develop a novel feature called blink similarity, which measures the similarity between the two blinking patterns. Our extensive experiments demonstrate that the proposed feature is quite robust, for it helps in Bell's Palsy detection even with very few labels. Our proposed eye-focused detection system is not only cheaper but also more convenient than several existing methods.\\
\end{abstract}

\copyrightnotice

\section{Introduction}
\label{sec:introduction}
Neurological disorders~\cite{singh2019predictive,yolcu2019facial} are diseases of the nervous system, which comprises the brain, spinal cord, and nerves. It is a highly complex system that coordinates all our actions and sensory information by transmitting signals through nerves to different parts of the body. Thus, a person suffering from a neurological disorder can have difficulty in bodily movements such as walking, speaking, swallowing, breathing, and blinking. These difficulties are often visible and can serve as excellent clues for diagnosing these disorders \cite{goyalgait}. Our goal is to build a computer-assisted diagnosis system for one such neurological disorder named Bell's Palsy. 

Bell's Palsy is a condition where muscles on one side of the face become weak or paralyzed. It affects 1 in every 60 persons at least once in a lifetime \cite{sullivan2016antiviral}. Moreover, according to \cite{pitts1988recurrent}, it gets repeated in 7\% of such patients. It makes one-half of the face appear to droop; the smile becomes one-sided, and eye muscles on the paralyzed side become weak and resist closing, causing the blink rate to go down. Currently, its diagnosis relies mainly on the visual inspection by a clinician. In an attempt to automate the diagnosis, this paper presents how just the blink-related clue just discussed can help build an effective Bell's Palsy diagnosis system.  

Existing computer vision algorithms can very easily and accurately locate the eyes to facilitate such study of blinks. Previously also, some disorders have been detected exclusively based on the eyes, such as computer vision syndrome \cite{divjak2009eye}, neurodegenerative disorders \cite{anderson2013eye}, Autism Spectrum Disorders \cite{10.1007/978-3-319-46723-8_37}, etc. However, to the best of our knowledge, no such eye-focused work has been proposed for Bell's Palsy so far. In Bell's Palsy\cite{wang2016automatic}, the drop in the blinking rate of the affected side causes a clear contrast between the blinking rates of the two sides. Our main idea is to model this contrast as a predictor (feature) for building the required system. 

However, there are some challenges in building such a system: (i) There are not many publicly available videos of the patients suffering from Bell's Palsy to develop an end-to-end data-driven model. Whatever model we build should be so robust that it works with even very few labels. (ii) There are no ready-made video-level features that capture the blink-related contrast discussed in the above paragraph. Although there are several works on images, there are very few prior works that deal with videos to solve this problem. (iii) Users may have privacy concerns too, and the system needs to have a privacy-preserving feature to facilitate the anonymity of the data collected. 

We propose a hybrid approach that efficiently blends deep learning and traditional machine learning (ML) approaches. We attempt Bell's Palsy detection (for which there is limited video data available) at the video level by first performing blink detection (for which there is plenty of image data available) at the frame level. Essentially, we break the problem into two stages: frame-level blink detection and video-level Bell's Palsy detection. While we solve the blink detection problem through the feature learning idea prevalent in deep learning, we rely on the hand-crafted feature design approach dominant in traditional machine learning for solving the Bell's Palsy detection problem. To solve these two problems, we develop two datasets by collecting images and videos from the internet. Our algorithm needs only eye regions. Hence, the subjects need not have privacy concerns because the entire face is no more needed.

This paper makes three contributions: (1) We develop novel benchmark datasets for detecting Bell's Palsy and blinks. (2) We create a novel eye-focused feature named blink similarity for videos through a hybrid approach. (3) We show that a Bell's Palsy diagnosis system can be build while incorporating privacy concerns by just focusing on the eyes.

\section{Related Work}
Since this work deals with blink detection and Bell's Palsy detection, we provide literature reviews of both in this section.  

The blink detection problem is a well-explored research problem. There are heuristic methods such as \cite{appel2016brightness}, where authors exploit the brightness feature. Then, there are traditional ML-based methods using HOG \cite{schillingmann2015yet} feature, motion \cite{anas2017online} feature, etc. With the rise of deep learning, researchers have explored several CNN-based approaches recently, such as \cite{anas2017online,kim2017study,cortacero2019rt}. We also use a CNN-based architecture for this problem. However, our main contribution to this problem is the dataset we provide, which is at the magnitude of nearly 40k images, which is the largest dataset built so far for this particular problem, to the best of our knowledge. Also, none of the existing works explicitly used blink detection to solve Bell's Palsy problem, to the best of our knowledge. 

 As far as Bell's Palsy is concerned, in \cite{park2010pc}, the authors analyze the video clips captured with webcam of PC for diagnosis by measuring the asymmetry index around the mouth region. In contrast, our work focuses on the eyes to detect Bell's Palsy. In \cite{wang2014automatic}, authors propose a method named ASMLBP (Active Shape Models plus Local Binary Patterns). In ASMLBP, the face is divided into eight local regions to describe each region using the facial points extracted using Active Shape Models (ASM) and through region descriptors of Local Binary Patterns(LBP). However, this method requires videos of the subjects to be taken in a controlled environment (see \cite{wang2014automatic} for more details). It also needs specific movements to be carried out for accurate detection. Similarly, \cite{he2009quantitative,ngo2016quantitative} also have constraints with video recording environment, video lengths, etc. Our proposed method does not have any such constraints.

In \cite{kim2015smartphone}, the authors propose a smartphone-based facial palsy diagnostic system that can work in the wild. They localize and track facial landmarks with a modified linear regression method and then compute the displacement ratio between the left and right side of facial landmarks located on the forehead and mouth region. Another work \cite{asthana2014incremental} proposes an incremental face alignment method to handle videos in the wild for this problem. However, there is always a possibility of facial points mismatching in such analysis. Recently, \cite{hsu2018deep} proposed a method to analyze facial Palsy using a deep hierarchical network by detecting asymmetric regions. Again, it depends upon the accuracy of facial landmarks detected. As far as the proposed method is concerned, we use facial landmarks only for localization \cite{10.1007/978-3-319-46478-7_12,8290832,7457899,8269367} of the eye regions; therefore, we can tolerate inaccuracy in landmarks detection up to a significant tolerance level. Most close to our work is \cite{storey20193dpalsynet}, which attempted facial palsy detection using videos by applying transfer learning on the 3D-CNN of \cite{hara2018can}. However, such end-to-end networks are hardly explainable. As a two-stage algorithm, our algorithm with an intuitive reason for intermediate detection of blinks has an edge in explainability.

\section{Proposed Method}

\begin{figure*}
	\begin{center}
		\includegraphics[width=1\linewidth]{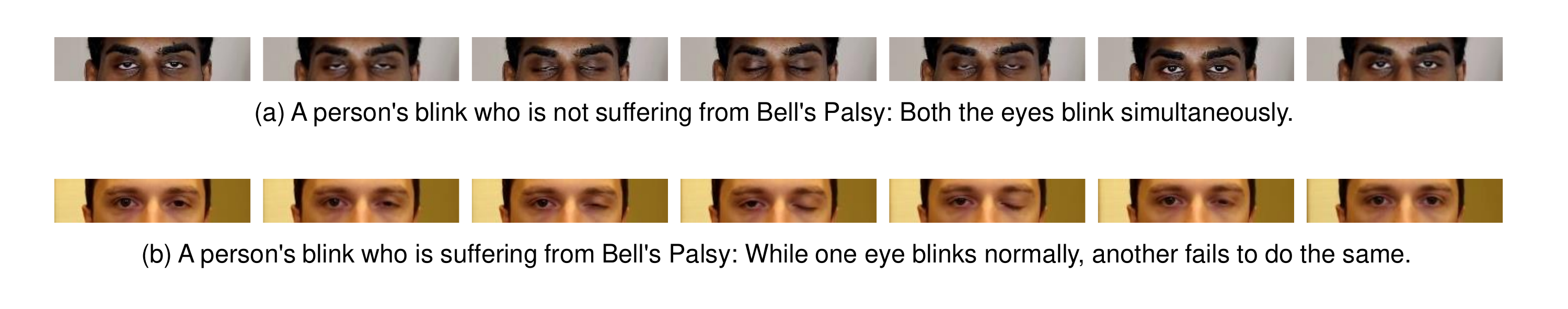}
	\end{center}
	\caption{This illustration demonstrates the difference between the blink of (a) a normal person and (b) a Bell's Palsy patient. While blinking patterns of the two eyes appear the same for a normal person, they differ a lot for a Bell's Palsy patient. Note how the patient's right eye remains open while his left blinks normally.}
	\label{fig:bpill}
\end{figure*}

Eye-blinking is a biological phenomenon of regular closure of eyelids for a brief period. As discussed earlier, in Bell's Palsy, one side of the face becomes weak or paralyzed, which seriously affects eye-blinking on that side. It leads to an abnormal contrast in the blinking patterns of the two eyes. As illustrated in Fig.\ref{fig:bpill}, there is a significant difference between the blinking patterns of (a) a normal person and (b) a Bell's Palsy patient. While both the eyes of a normal person blink together, only the patient's normal side's eye blinks. Hence, we can exploit this observation related to the eye-blinks for detecting Bell's Palsy in a person. However, even normal persons may sometimes blink just one eye when they wink or when an instantaneous agitation occurs. Therefore, the blinks need to be monitored for a substantial time, necessitating a video-based diagnosis system.

Let $ECT$ denote the total time duration during which an eye remains closed due to the blinking phenomenon in a video. For a patient, the $ECT$ of the paralyzed side will be much lower than that of the normal side. On the contrary, for a normal person, both the $ECT$s should be almost equal.          

There are two phases in the eye-blinking activity: eye-closure time ($EC$) and eye-open time ($EO$). If we assume eye-blinking to be a periodic activity, these phases tend to have constant values for a particular eye, although they may vary eye-to-eye and person-to-person. Thus, the blink time-period becomes $EC+EO$. If $L$ denotes the length (total time duration) of a video, we can now compute $ECT$ for an eye as
\begin{equation}\label{e1}
ECT=\frac{EC}{EC+EO}\times L,
\end{equation} 
where ($\frac{EC}{EC+EO}$) serves as fraction coefficient of $L$ to compute $ECT$. So, multiplying the length of video with such a fraction yields the total time duration during which the eye remains closed, i.e., $ECT$. Although we can extract $L$ from the meta-data of a video, $EC$ and $EO$ are variables not only across different persons but also possibly between two eyes of the same person (especially, patient). Therefore, they are difficult to compute. Nevertheless, we can also compute the number of eye-closed frames, denoted as $ECF$, using the same fraction coefficient, i.e.,
\begin{equation}\label{e2}
ECF=\frac{EC}{EC+EO}\times F,
\end{equation} 
where $F$ is the total number of frames. By dividing Eqn.\eqref{e1} with Eqn.\eqref{e2}, we get a relationship between $ECT$ and $ECF$, which is as follows:
\begin{equation}
ECT = ECF\times\frac{L}{F}.
\end{equation} 

Considering both the left ($l$) eye and the right ($r$) eye now, we have $ECT_l=ECF_l\times\frac{L}{F}$ and $ECT_r=ECF_r\times\frac{L}{F}$. By dividing the two relationships with one another, we can conclude that 
\begin{equation}
\frac{ECT_l}{ECT_r}=\frac{ECF_l}{ECF_r}.
\end{equation}

Thus, the ratio of two $ECT$s is equal to the ratio of two $ECF$s, and there is no dependence on video parameters like $L$ and $F$ to compute that ratio. In this way, we model a ratio of physical parameters ($ECT$s) as a ratio of computable parameters ($ECF$s). Therefore, we need only to find the number of frames in which the left eye is closed ($ECF_l$) and the number of frames in which the right eye is closed ($ECF_r$) to compute the ratio. This ratio should be close to 1 for normal persons and deviate from 1 for Bell's Palsy patients.  

\begin{figure*}
	\begin{center}
		\includegraphics[width=1\linewidth]{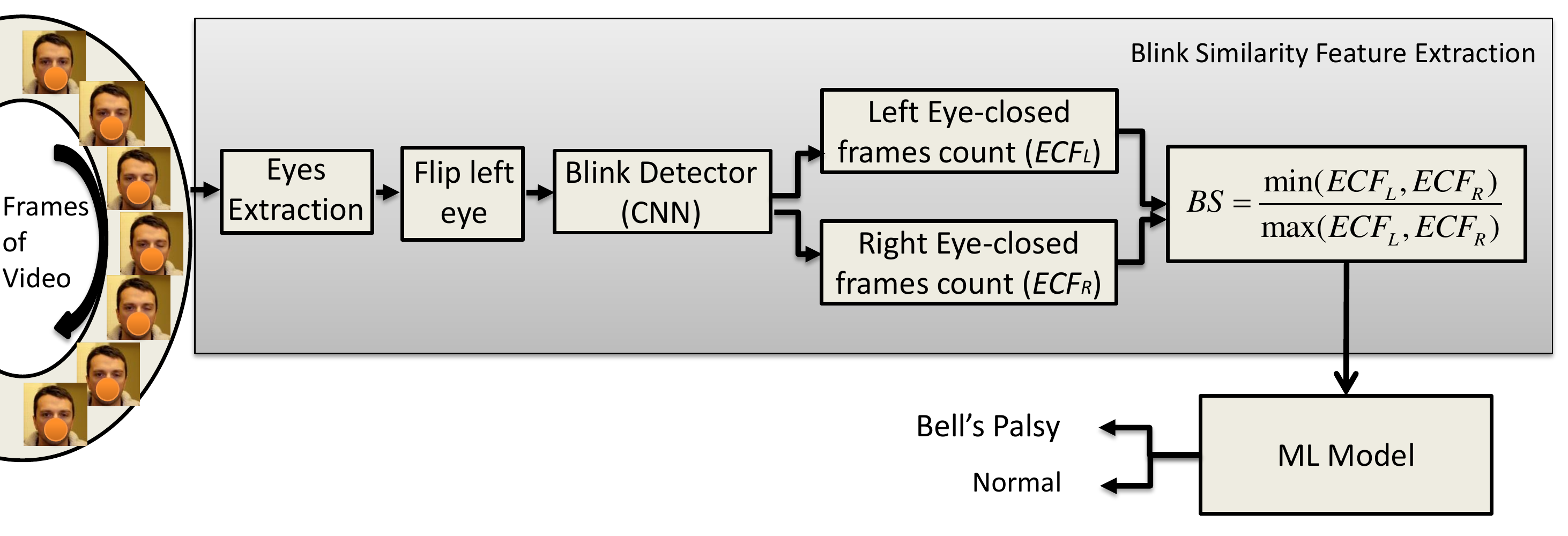} 
	\end{center}
	\caption{Proposed approach for the extraction of blink-rate similarity feature and the end-to-end workflow for the detection of Bell's Palsy.}
	\label{fig:bp}
\end{figure*}

\subsection{Blink Similarity}

Inspired by the above ratio, which can identify Bell's Palsy patients, we modify it so that it has a physical meaning. We propose a novel video-level feature called blink similarity ($BS$), which we compute in the following manner:
\begin{equation}
BS=\frac{min\big(ECF_l,ECF_r\big)}{max\big(ECF_l, ECF_r\big)},
\end{equation}
where the ratio automatically becomes $\frac{ECF_l}{ECF_r}$ or $\frac{ECF_r}{ECF_l}$ depending upon whether the paralyzed side is the left one or the right one, respectively, because the eye-closure time of the eye in the paralyzed side will be less than that of the other eye. More than that, it has a physical meaning of similarity of blinks. It varies from 0 to 1 such that the more the value more is the similarity of the two blinking patterns.

To develop this feature, we need an intermediate detection model at frames level that detects blinks (eye-closure). Also, to effectively use this feature, we need a final detection model at the video level that detects Bell's Palsy. We will discuss both these models next. We give the entire end-to-end pipeline in Fig.~\ref{fig:bp}.

\begin{figure}
	\begin{center}
		\includegraphics[width=1\linewidth]{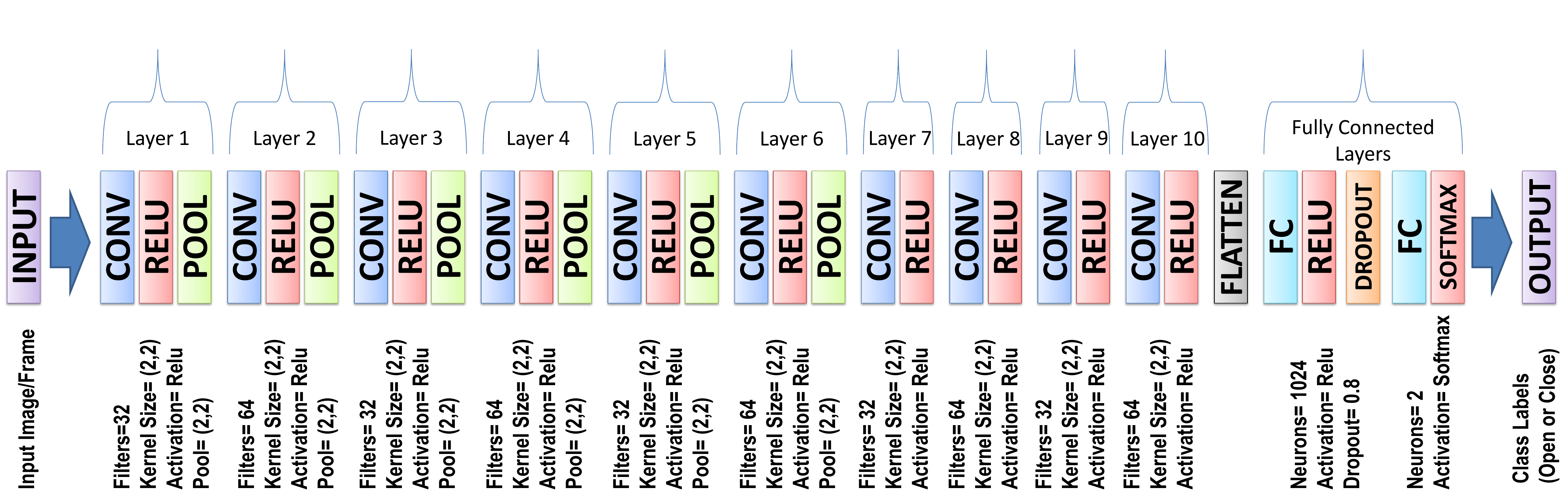}
	\end{center}
	\caption{CNN architecture used for developing our intermediate detection model (Blink Detector).}
	\label{fig:cnn}
\end{figure}

\subsection{Intermediate Detection Model}

To detect the closed eye, we build an intermediate detector named blink detector using a Convolutional Neural Network (CNN). It takes eye images as input and outputs the state of the eye, i.e., whether it is closed or open. From any frame, we extract the person's two eyes and flip the left one sideways because we train the CNN on all right-types (by flipping the left-types). For creating the blink detector, we build a blink detection dataset (details in the next section) comprising open and closed eye images to train a convolutional neural network (CNN). As presented in Fig.~\ref{fig:cnn}, the architecture's convolutional base consists of 10 convolutional layers, each followed by ReLU (Rectified Linear Unit) and Max-pooling layers. The convolutional base's output is flattened and fed to the fully connected layers to generate the final output. The last fully connected layer uses the softmax activation function, while the other fully connected layer uses ReLU as its activation function. The purpose behind having an additional Dropout (0.8) layer is to prevent overfitting. The input images for the blink detector are of resolutions 50x50. Our CNN uses the categorical cross-entropy loss ($CEL$), as mentioned below: 

\begin{equation}
CEL(y,\hat y{}) =-\sum_{j=1}^{C} \sum_{i=1}^{|I|} \Big(y_{ij}*\log(\hat{y}_{ij})\Big)
\end{equation}
where $y_{ij}$ is the ground-truth label value of $i^{th}$ image with respect to $j^{th}$ class, and $\hat{y}_{ij}$ is the predicted label value of $i^{th}$ image with respect to $j^{th}$ class. $C$ is the number of classes, which is two (2) in this case. $|I|$ is the total number of images in a mini-batch. We use the Adam optimization algorithm to train this CNN.

\subsection{Final Detection Model}

Since we now have a video-level feature in the form of a blink similarity feature, we can tune over several supervised learning algorithms to build the best final detection model based on cross-validation classification accuracy. We found that Bell's Palsy is best-detected using a model developed by applying Stochastic Gradient Descent (SGD) with hinge loss. The model thus developed is a binary classifier that detects if the subject present in the video has Bell's Palsy or not by simply utilizing the intuitive blink similarity feature developed.  

\begin{figure*}
	\begin{center}
		\includegraphics[width=1\linewidth]{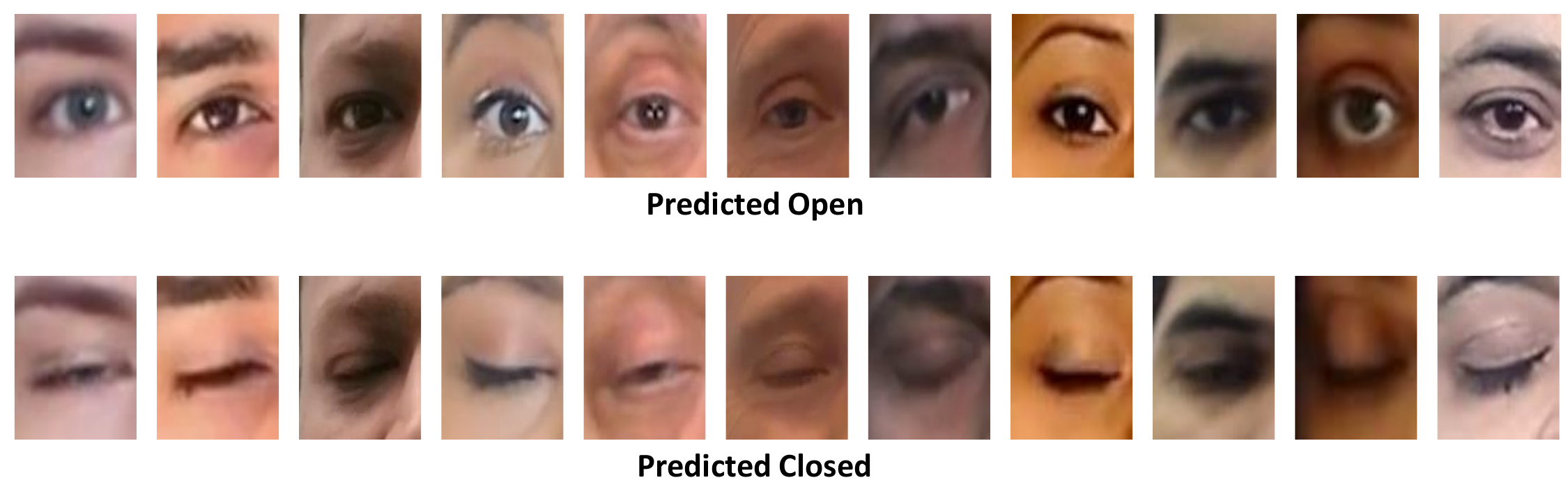}
	\end{center}
	\caption{Sample results of our blink detector.}
	\label{fig:bdr}
\end{figure*}

\section{Experimental Results}
In this section, we discuss the datasets created, the evaluation process employed, our detectors' performances, and various comparisons. 

\subsection{Datasets}

We develop two datasets: (i) One for developing the intermediate detection model (Blink Detector). (ii) Another one for developing the final detection model (Bell's palsy detector). While the intermediate one takes image/frame as input, the final one takes our video-level feature as inputs. While the intermediate detector's output is whether there is a blink or not, the final detectors' output is whether the person has Bell's Palsy or not. 

\textbf{Blink Detection Dataset}: It consists of 21896 open eye images and 19753 closed eyes images. The purpose is to detect blinks, i.e., to detect if a given eye image contains a closed eye. Most of these images were collected from YouTube videos by manually cropping out the eyes frame-by-frame.

\textbf{ Bell's Palsy Video Dataset}: It consists of 41 videos of patients suffering from Bell's Palsy and 34 videos of normal persons. We collect most of these videos from YouTube.  

We also test our method on publicly available YouTube Facial Palsy(YFP) \cite{hsu2018deep} dataset. The YFP dataset contains videos of only Bell's Palsy patients.

\begin{table*}
	\begin{center}
		\caption{Comparison of our Intermediate detector with those built using VGG and Inception-ResNet features on our Blink Detection Dataset.} \label{tab:id}
		\begin{tabular}{|l|l|l|l|}
			\hline
			Detectors &Training&Validation&Testing\\
			\hline
			\hline
			VGG + Random Forest Blink Detector&\textbf{99.96}\%&98.28\%&98.09\%\\
			Inception-ResNet + Random Forest Blink Detector&99.90\%&97.80\%    &97.56\%\\
			\hline
			Proposed Blink Detector&99.80\%&\textbf{99.72}\%&\textbf{99.57}\%\\
			\hline
		\end{tabular}
	\end{center}
\end{table*}

\begin{table}
	\begin{center}
		\caption{Confusion matrix obtained during the testing phase while building our intermediate detector on our blink detection dataset.} \label{tab:cmid}
		\begin{tabular}{|l|l|l|l|}
			\hline
			\multirow{4}{*}{Blink Detector}& &Predicted&Predicted\\
			& &Open&Close\\
			\cline{2-4}
			&Actually Open& 3538&15\\
			&Actually Close& 7&2849\\
			\hline
			
		\end{tabular}
	\end{center}
\end{table}

\begin{table*}
	\begin{center}
		\caption{Comparison of 3DPalsyNet \cite{storey20193dpalsynet} features with our Blink Similarity feature in terms of classification accuracies obtained during cross-validation using different machine learning algorithms on our Bell's Palsy Video dataset.} \label{tab:bpc}
		\begin{tabular}{|l|l|l|l|l|l|l|l|l|l||l|}
			\hline
			&AB&LR&NB&NN&RF&SGD&SVM&DT&kNN&Average\\
			\hline
			3DPalsyNet \cite{storey20193dpalsynet}&62.2&73.8&66.8&81.6&68.2&71.0&70.8&63.0&79.7&70.8\\
			\hline
			Blink Similarity (Ours)&\textbf{90.7}&\textbf{92.0}&\textbf{90.7}&\textbf{93.3}&\textbf{93.3}&\textbf{94.7}&\textbf{93.3}&\textbf{90.7}&\textbf{93.3}&\textbf{92.4}\\
			\hline
		\end{tabular}
	\end{center}
\end{table*}

\begin{table}
	\begin{center}
		\caption{Confusion Matrix obtained during cross-validation of our final detector on our Bell's Palsy Video dataset.} \label{tab:cmifd}
		\begin{tabular}{|l|l|l|l|}
			\hline
			\multirow{3}{*}{Bell's Palsy}&&Predicted No&Predicted Yes\\\cline{2-4}
			&Actual No& 32&2\\
			&Actual Yes& 2&39\\
			\hline
		
		\end{tabular}
	\end{center}
\end{table}

\subsection{Evaluation}

We report classification accuracy, which means the percentage of correct predictions. We take different evaluation approaches for intermediate and final detections given the disparity in the labeled data available for the two problems.    

For intermediate detector, since we have labeled data in abundance, we can divide the labeled data into three parts: training, validation, and testing. Specifically, we use the hold-out approach to validation. In the hold-out approach, a portion of the dataset is kept aside for validation. We manage to set the best hyperparameters while training the CNNs for building the intermediate detector with the help of the validation scores. Once we set them, we test the trained CNN on the testing dataset.    

In contrast, since we have minimal labeled data for the final detector, we avoid the testing phase altogether and employ the k-fold cross-validation approach on the entire labeled dataset. Note that we set k to 3. The cross-validation helps us in identifying the best algorithm to use among nine basic machine learning algorithms, namely AdaBoost (AB), Logistic Regression (LR), Naive Bayes (NB), Neural Networks (NN), Random Forests (RF), Stochastic Gradient Descent (SGD), Support Vectors Machine (SVM), Decision Tree (DT) and k-Nearest Neighbors (kNN). Once we choose the best learning algorithm, we train our model on the entire labeled dataset to create our final detectors. We use the default setting (as per Python's Orange package) of hyper-parameters for these algorithms. In addition to the classification accuracy, we also report appropriate confusion matrices while reporting our detectors' performances.

\subsection{Intermediate Detection Results}

We give the quantitative results of our intermediate detector in terms of classification accuracy in Table~\ref{tab:id}, while also comparing with other detectors built using Random Forest upon the features extracted from the pre-trained deep learning models like VGG19 and Inception-ResNet. Our detector obtains superior results compared to others in challenging validation and testing phases. It obtains classification accuracy of above $99\%$ in all three phases. We split the datasets into three parts: 70\% for the training, 15\% for validation, and 15\% for testing. The confusion matrix obtained during the testing phase using our intermediate detector is given in Table~\ref{tab:cmid}. We give the sample detection results of our intermediate detector in Figure~\ref{fig:bdr}. Our blink detector detects very well whether the eye is in the open state or closed state.

\subsection{Final Detection Results}
We build our final detector using the blink similarity feature we have developed at the video level. The quantitative results of our final detector in terms of classification accuracy during cross-validation are given in Table~\ref{tab:bpc}. It is clear from the table that SGD (Stochastic Gradient Descent) is the best algorithm for detecting Bell's Palsy using our feature. Even with very few labels, the algorithms' high average accuracy shows that our feature is robust, thanks to our disorder-specific design. Like the intermediate detector, for the final detector also, we provide the cross-validation confusion matrices in Table~\ref{tab:cmifd}. It clearly shows that most videos are being correctly classified. Only four videos are misclassified.           
\subsection{Comparisons}
There is a publicly available video dataset name YouTube Facial Palsy (YFP) dataset. It consists of 31 videos of Bell's Palsy patients. On this dataset, we test the model trained on our Bell's Palsy Video dataset. It could detect 28/31 videos as Bell's Palsy videos, showing our method can work cross-domain too. Since there is also a prior work\cite{storey20193dpalsynet} on detecting Bell's Palsy from videos, we extract the features from the network presented in \cite{storey20193dpalsynet} and perform the same experiments as we did with our Blink Similarity feature on our Bell's Palsy Video dataset. The results are compared with ours in Table~\ref{tab:bpc}. It can be seen that our blink similarity feature outperforms features of \cite{storey20193dpalsynet} no matter which machine learning algorithm is used. \cite{storey20193dpalsynet} achieves the best cross-validation accuracy of 81.6\% using NN, whereas our feature achieves the best cross-validation accuracy of 94.7\% using SGD. In terms of average accuracy across the machine learning algorithms, while \cite{storey20193dpalsynet} obtains 70.8\%, our feature obtains 92.4\%. This comparison shows that our blink similarity feature is superior to that of \cite{storey20193dpalsynet}, thanks to our disorder-focused design.
\subsection{Discussion \& Limitations}
Our feature can also provide the severity degree of the disorder. The farther the blink similarity feature value from 1, the more severer is the Bell's Palsy. Thus, the proposed feature can also assist in the rehabilitation \cite{avola2018vrheab} process.  
Note that our proposed method entirely relies on detecting blinks. It might be possible that such events may not get adequately captured in some videos. For example, there could also be a short video that not even captures one blink. Similarly, there could be a video that captures only one eye. In both cases, our method would fail. These are some of the limitations of our proposed method.
\section*{Conclusion}
We attempted a novel problem called eyes-focused detection of Bell's Palsy. We design a novel eye-based feature named blink similarity at the video level. The feature is so robust that it helps build a machine learning model with even very few labels. While designing the feature, we also had to develop an intermediate detection model (a CNN) for blink detection. In this way, we propose a two-stage hybrid approach while efficiently leveraging the benefits of both the feature learning and hand-crafted designing of the features. We build two datasets: one for blink detection and another for Bell's Palsy detection. Our extensive experiments demonstrate better performance than existing methods.


\printbibliography[heading=subbibintoc]

\end{document}